# *k*-Relevance Vectors: Considering Relevancy Beside Nearness

Sara Hosseinzadeh Kassani[1], Farhood Rismanchian[2], Peyman Hosseinzadeh Kassani[3,*]

[1]Department of Computer Science, University of Saskatchewan, Saskatchewan, Canada

[2] Department of Health Services Management, Isfahan University of Medica Sciences, Isfahan, Iran

[3]Neurology and Neurological Department, Stanford University, Palo Alto, CA, USA

## Highlights

- New concept of similarity; relevancy beside nearness.
- Proposed a highly sparse method with good classification accuracy.
- Proposed a new parameter that helps to improve the final accuracy.
- Validated the proposed model on several datasets.

**Abstract:**

This study combines two different learning paradigms, k-nearest neighbor (k-NN) rule, as memory-based learning paradigm and relevance vector machines (RVM), as statistical learning paradigm. The purpose is to improve the performance of k-NN rule through selection of important features with sparse Bayesian learning method. This combination is performed in kernel space and is called k-relevance vector (k-RV). The proposed model significantly prunes irrelevant features. Our combination of k-NN and RVM presents a new concept of similarity measurement for k-NN rule, we call it k-relevancy which aims to consider "relevancy" in the feature space beside "nearness" in the input space. We also introduce a new parameter, responsible for early stopping of iterations in RVM that is able to improve the classification accuracy. Intensive experiments are conducted on several classification datasets from University of California Irvine (UCI) repository and two real datasets from computer vision domain. The performance of k-RV is highly competitive compared to a few state-of-the-arts in terms of classification accuracy.

## 1- Introduction

Pattern classification is one of the major components of intelligent systems and has numerous applications in several research areas [1-7]. The purpose is to assign unseen / test data to either the positive or negative category [8]. Support vector machines (SVM) [9], decision trees [10], *k*-nearest neighbor (*k*-NN) rule [11], naïve Bayes [8] are most popular classifiers. Among them, the simple yet powerful *k*-NN rule is one of the top ten known data mining algorithms [12] and has attained popularity since 1967 [11]. *k*-NN rule is an instance-based learning algorithm in which the prediction for a query instance **z** is regionally calculated through nearest neighbors of **z**. Nearness is known as similarity measurement and is usually calculated with Euclidean distance between **z** and its neighbors. If number of neighbors, *k* is equal to 1, **z** is assigned to the class of the nearest neighbor, i.e. the one has shortest Euclidean distance to the **z**. The complexity of *k*-NN rule is O(*Nl*) where *N* is the total number of training data and *l* is the size of data columns. This complexity increases with the value of *k* as the number of instances. Additionally, different values of *k* may alter the target value of the query instance **z**, and hence, appropriate selection of *k* is a non-trivial task.

*k*-NN has been applied to several applications such as medical diagnosis [13-15], intrusion detection [16, 17], image classification [18-20], video multimedia classification and segmentation [5, 21, 22]. Regardless of the merits of *k*-NN rule, it has two problems in learning from data. *First*, *k*-NN uses distance metric to calculate similarities between instances. *k*-NN assumes all variables equally contribute to the learning procedure which may lead to poor prediction in the presence of redundant/noisy variables. Additionally, the main patterns in data are contained in few variables. In this sense, weight assignment may better increase the accuracy of the nearest neighbor rule [23]. *Second*, the presence of noise in data may jeopardize the *k*-NN performance as it is necessary to calculate the similarity between instances through distance metrics like Euclidean metric [23]. For this problem, one can modify the distance measurement. The feature importance scores can be evaluated via weight assignment to each feature. This method is known as feature weighting method [23-28].

Tutz and Ramzan [28] used *w*-NN for imputation of missing data. The careful selection of distances with weight's assignment helped discover more information of the missing values and outperformed competing nearest neighbor methods.

The main idea in [23] is to refine the distance metric of *k*-NN based on the Minkowski distance in order to include only a subset of relevant features. Experimental results on low- and high-dimensional datasets demonstrate the importance of these modifications. Takeuchi et.al propose a simultaneous feature selection and weighting method [25] applied to *k*-NN classifier. With a novel feature weighting algorithm, the correct target neighbor is obtained, and nearest neighbors are updated through a sequential quadratic programming. This approach gained better performance than traditional *k*-NN.

In our study, using a simple yet versatile algorithm, the proposed *k*-RV rule may less suffer from same problems as remarked in above. We take advantage of sparse Bayesian learning (SBL) approach in RVM to prune irrelevant and noisy features [29-31]. The sparse weight vector obtained by RVM is used as weight values for each variable in *k*-NN rule.

In [32], two drawbacks of extreme learning machine (ELM) [33], i.e. 1) suffering from overfitting problem and 2) sensitivity of the accuracy of ELM to the number of hidden neurons are resolved by combining ELM with SBL approach in RVM [31]. During the learning process, several redundant hidden neurons of ELM are detected and pruned.

Study in [34], also reduces the number of hidden neurons of multivariate polynomial (MP) neural network and reduced polynomial (RP) [35] with the sparsity approach in RVM. The sparsed methods, i.e. sparse Bayesian RP (SBRP) and sparse Bayesian MP (SBMP) have better classification accuracy than MP and RP evaluated on several datasets.

In a similar fashion to [32, 34] as the source of inspiration for our study, we also modify the distance metric in *k*-NN rule through SBL method used in RVM in order to obtain a compact set of features that are more relevant to predictor variable **t**. The learned sparsed weights obtained by RVM are served as feature scores and are fed into the distance metric of *k*-NN. And hence this builds up a sparsed *k*-NN rule, that we henceafter call *k*-relevance vector (*k*-RV). Our proposed *k*-RV adds a novel concept of similarity measurement into *k*-NN rule, we call *k*-relevancy which aims to consider "*relevancy*" beside "*nearness*".

The contributions and advantages of *k*-RV are as follows:

1. We link two different learning approaches, *k*-NN as memory-based learner and SBL as statistical method for feature selection. Our combination adds a novel concept of similarity measurement into *k*-NN, we call *k*-relevancy, i.e. considering *relevancy* than *nearness*.
2. We introduce a new parameter for early stopping of iterations in RVM. If query instance is far away from the RVs, then RVM may have poor prediction for that query instance [37]. We abort iteration in RVM if the change of $\alpha$ (The variable $\alpha$ in RVM is responsible for pruning training data) between current iteration *i* and previous iteration *i*-1 is smaller than a threshold value. We gain higher performance of *k*-RV through tuning this new parameter.
3. Intensive experiments on a generated toy data, several UCI datasets and two real-world data from computer vision domain validate the reliability of *k*-RV compared with several counterparts.

The remainder of this paper is organized as follows: Section two covers the related works, motivation and contributions. Section three discusses *k*-NN rule and RVM as preliminaries. Section four expresses the proposed *k*-RV rule. Some theoretical points are explained in discussion section. Section six examines experiments conducted on several UCI datasets and two real databases chosen from computer vision domain. Last section pays attention to remarks and concludes our work.

## 2. Preliminaries
### 2.1. *k*-NN and *w*-NN rules

Assume $(\mathbf{x}_1, \mathbf{t}_1), (\mathbf{x}_2, \mathbf{t}_2), \ldots, (\mathbf{x}_N, \mathbf{t}_N)$ be $N$ data points in which $\mathbf{x}_i$ is the training input and with $D$ dimensions and $\mathbf{t}_i$ is the target value. To find the $k$ nearest neighbors of the query instance $\mathbf{z}$, the Euclidean distance is performed as follows:

$$Dist_{k-\text{NN}}(\mathbf{x}, \mathbf{z}) = \sqrt{\sum_{i=1}^{D}(x_i - z_i)^2} \qquad (1)$$

where **x** and **z** are respectively train and test vectors. The pairwise distance comparison between the query instance **z** and each of *N* training instances **x** is computed. The *k* number of training instances with shortest distance to **z** are nearest neighbors.

Moreover, the importance of each attribute can be evaluated through weight assignment into Eq. (1). This method is known as weighting nearest neighbor rule or *w*-NN. The new equation for *w*-NN is as follows:

$$Dist_{w-\text{NN}}(\mathbf{x}, \mathbf{z}) = \sqrt{\sum_{i=1}^{D} w_i (x_i - z_i)^2} \qquad (2)$$

where weight $w_i$ for each attribute may change the order of neighbors and may improve the accuracy of *k*-NN.

## 2.2. Relevance vector machine

Sparse Bayesian learning (SBL) and in particular relevance vector machine (RVM) [38] is a probabilistic formulation to support vector machine (SVM) [39]. To estimate the target vector $\mathbf{t} \in \mathbb{R}^{1 \times N}$ in kernel space, we suppose $\mathbf{t} = \mathbf{w}^T \mathbf{H}$ where $\mathbf{w} \in \mathbb{R}^{(N+1) \times 1}$ is weight vector and $\mathbf{H} \in \mathbb{R}^{(N+1) \times N}$ is design matrix with $N$ training instances. For classification task, each training instance $\mathbf{x} \in \mathbb{R}^{N \times l}$ with $l$ dimension is supposed to be drawn from an independent Bernoulli random variable with probability $p(\mathbf{t}|\mathbf{x})$ where $\mathbf{t} \in \mathbb{R}^{1 \times N}$ is a target vector. Additionally, one can choose posterior probability $p(\mathbf{w}, \boldsymbol{\alpha}|\mathbf{t})$ as objective function with $\boldsymbol{\alpha} \in \mathbb{R}^{(N+1) \times 1}$ as hyperparameter vector. Showing this for $N$ training instances with Bernoulli likelihood,

$$p(\mathbf{t}|\mathbf{w}) = \prod_{i=1}^{N} \sigma\{g(\mathbf{h}_i; \mathbf{w})\}^{t_i}[1 - \sigma\{g(\mathbf{h}_i; \mathbf{w})\}]^{1-t_i} \quad (3)$$

where $\mathbf{k}_i$, $g$, and $\sigma$ are respectively $i^{th}$ column vector of the kernel matrix $\mathbf{H}$, the network output and the nonlinear Sigmoid function. A closed form solution cannot be achieved since Sigmoid is a nonlinear function. Hence, according to [38], iterative Mackay procedure [40] is necessary to get the marginal likelihood.

To solve the posterior probability $p(\mathbf{w}, \boldsymbol{\alpha}|\mathbf{t})$ in an easy way, its decomposed part, i.e. marginal likelihood $p(\mathbf{w}|\mathbf{t}, \boldsymbol{\alpha})$ is optimized:

$$p(\mathbf{w}|\mathbf{t}, \boldsymbol{\alpha}) \propto p(\mathbf{t}|\mathbf{w})p(\mathbf{w}|\boldsymbol{\alpha}) \quad (4)$$

Let $p(w_i|\alpha_i) \approx N(0, \alpha_i^{-1})$, and then three subsequent techniques, namely marginalization, Laplace approximation, and Iterative Regularized Least Square (IRLS) are used to calculate $\widehat{\mathbf{w}} \in \mathbb{R}^{(N+1) \times 1}$ and $\Sigma \in \mathbb{R}^{(N+1) \times (N+1)}$ as follows:

$$\widehat{\mathbf{w}} = \Sigma \mathbf{H}^T \mathbf{B} \mathbf{t} \quad (5)$$

$$\Sigma = (\mathbf{H}^T \mathbf{B} \mathbf{H} + \mathbf{A})^{-1} \quad (6)$$

where $\widehat{\mathbf{w}} \in \mathbb{R}^{(N+1) \times 1}$ is posterior mode of $\mathbf{w}$ and $\mathbf{A} = diag(\boldsymbol{\alpha})$ and $\mathbf{B}$ is a diagonal matrix with diagonal elements $(\beta_1, \beta_2, \dots, \beta_N)$ where $\beta = \sigma\{t(\mathbf{x}_n)\}[1 - \sigma\{t(\mathbf{x}_n)\}]$. Once $\Sigma$ and $\widehat{\mathbf{w}}$ are initialized, the hyper-parameters $\alpha_i$ are updated as follows,

$$\alpha_i^{new} = \frac{1 - \alpha_i \Sigma_{ii}}{\widehat{w}_i^2} \quad (7)$$

where $\widehat{w}_i$, and $\Sigma_{ii}$ are the $i^{th}$ posterior weight and the $i^{th}$ diagonal element of $\mathbf{\Sigma}$ computed with Eq. (5) and Eq. (6). This is an iterative algorithm and the parameters in Eq. (5), (6), and (7) should be repeated till to reach a stopping criterion. Since $\alpha$ has inverse relationship to $w$, many $\alpha$'s tend to infinity and hence the corresponding $w_i$'s tend to zero. Simply, pruning elements from vectors $\mathbf{w}$, $\boldsymbol{\alpha}$ and the corresponding basis vectors from $\mathbf{H}$ yields sparsity.

## 3. Proposed *k*-RV rule

### 3.1 Overview

The proposed *k*-RV is tailored to three main steps, namely, data kernelization with Gaussian and polynomial kernels, kernel data sparsification with RVM, and classification with sparsed weights. **Fig**. 1 shows workflow of the proposed *k*-RV.

**1- Data Kernelization (Feature expansion)**

**2- Kernel Data Sparsification with Bayesian method**

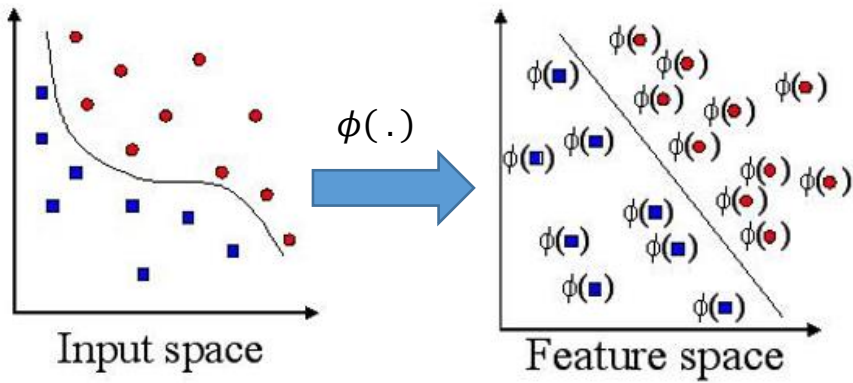

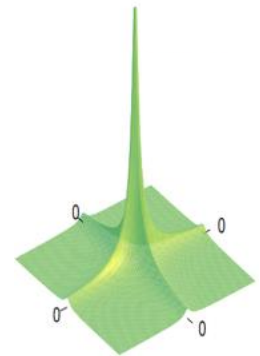

$\alpha \to \infty$ then $w \to 0$

**3- k-NN classification in Relevance space**

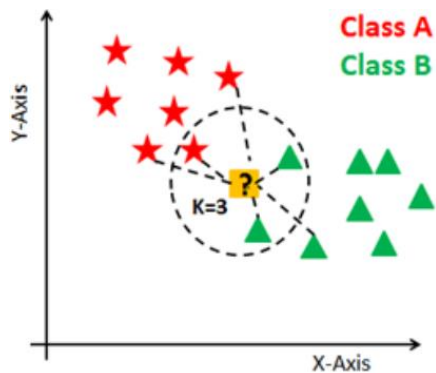

**Fig. 1**. Workflow of the proposed method; 1) Data kernelization: data is expanded to feature space that may help to make a linearly non-separable data to be separable. And, this step is necessary in making data sparse with sparse Bayesian learning in the next step; 2) Kernel data sparsification: those irrelevant or less informative features are pruned with the idea of sparse Bayesian learning in RVM; 3) *k*-NN classification with relevance neighbors: a k-NN is applied to sparsed data. As many features are pruned by the previous step, the distance of neighbors for every query instance z change and more likely is relevant to *z*.

### 3.2. *k*-RV: combination of *k*-NN and RVM

After kernelizing and pruning kernel matrix **H** with sparsity approach in RVM, the sparsed weights $\mathbf{w}^{sp}$ and corresponding sparsed kernel matrix $\mathbf{H}^{sp}$ are fed into the distance metric in nearest neighbor rule. Hence, the distance metric in *k*-RV is modified as follows:

$$Dist_{k-\mathrm{RV}}(\mathbf{h}^{sp}, \mathbf{z}^{sp}) = \sqrt{\sum_{i=1}^{D} w_i^{sp}\left(h_i^{sp} - z_i^{sp}\right)^2} \qquad (8)$$

where $z_i^{sp}$ is the *i*-th sparsed query instance. The class assignment for the test instance is performed with majority voting technique. The flowchart of the proposed *k*-RV is shown as follows:

---
**Algorithm 1. *k*-RV**

**Initialization and kernelization**
1. Initialize **w**, $\boldsymbol{\alpha}$, $\boldsymbol{\Sigma}$, and length of Gaussian kernel and the number of neighbors *k*.
2. Expand the input training matrix **X** to the feature matrix **H** with Gaussian or Polynomial kernel function.

**Sparsification**
  **While** $\Delta\alpha > 0.1$
      3. Find $\boldsymbol{\alpha}$ through $\alpha_i^{new} = \frac{1-\alpha_i \Sigma_{ii}}{\widehat{w}_i^2}$ with initialized **w** and $\boldsymbol{\Sigma}$.
      4. $\widehat{\mathbf{w}} = \boldsymbol{\Sigma}\mathbf{H}^{\mathrm{T}}\mathbf{Bt}$, $\boldsymbol{\Sigma} = (\mathbf{H}^{\mathrm{T}}\mathbf{B}\mathbf{H} + \mathbf{A})^{-1}$, $\alpha_i^{new} = \frac{1-\alpha_i \Sigma_{ii}}{\widehat{w}_i^2}$
  **End While**

**Classification**
  **For** *i* = 1: # test instances
      5. Calculate $Dist_{k-\mathrm{RV}}(\mathbf{h}^{sp}, \mathbf{z}^{sp}) = \sqrt{\sum_{i=1}^{D} w_i^{sp}\left(h_i^{sp} - z_i^{sp}\right)^2}$
      6. Assign the target value for every test instance using majority voting and RVs
  **End For**.

---

## 4. Experiments

### 4.1. Datasets and Setup

Twenty datasets are selected from UCI data repository. To select these datasets, we care about both small size and large size in terms of the number of instances, binary and multi-class classification tasks. The descriptions of datasets are listed in the **Table** 1.

Some notes for the experimental setup:

1. The performance of the proposed *k*-RV is compared with ker-NN, RVM with Gaussian likelihood (RVM-Gauss), and RVM with Bernoulli likelihood (RVM-Bern).
2. For *k*-NN based learners, the value for *k* is in the range $\{1, 2, 3, \ldots, 51\}$. To report the CA, we use the best value for *k* to predict the test labels. To find the best value for *k*, 10 runs of 10-fold cross validation is performed, and the average of CA is taken into consideration.
3. The new parameter $\Delta\alpha$ of *k*-RV is the difference between the $\boldsymbol{\alpha}$ at current iteration with the $\boldsymbol{\alpha}$ in the previous iteration. So, $\Delta\alpha$ is set to $\{10^{-6}, 10^{-5}, 10^{-4}, 10^{-3}, 10^{-2}, 10^{-1}, 1, 10\}$. Different $\Delta\alpha$ gives different number of RVs.
4. The width σ of Gaussian kernel $K(\mathbf{u}, \mathbf{v}) = \exp\left(\frac{\|\mathbf{u}-\mathbf{v}\|^2}{2\sigma^2}\right)$ for both ker-NN and *k*-RV is in the range of $\{0.05, 0.10, \ldots, 1\}$. The order *c* of polynomial kernel $K(\mathbf{u}, \mathbf{v}) = (\mathbf{u}.\mathbf{v} + 1)^c$ is set to $c = 2$ where **u** and **v** are two vectors in the input space.
5. The proposed model is implemented on a personal computer, with 32 GB RAM and with CPU 3.4 GHz, Intel Core i7. We use Matlab to do experiments.

**Table 1**. Descriptions of datasets*

| # | Dataset | # instances | # attributes (R/I/N)* | # Classes |
|---|---------|-------------|----------------------|-----------|
| Binary class problems ||||||
| 1 | Wbcd | 683 | 9 (0/9/0) | 2 |
| 2 | Australia | 690 | 33 (32/1/0) | 2 |
| 3 | Heart | 270 | 13 (1/12/0) | 2 |
| 4 | Teaching | 151 | 6 (0/6/0) | 2 |
| 5 | Ionosphere | 351 | 33 (32/1/0) | 2 |
| 6 | Pima | 768 | 8 (8/0/0) | 2 |
| 7 | Bupa | 345 | 6 (1/5/0) | 2 |
| 8 | Shuttle | 253 | 7 (0/7/0) | 2 |
| 9 | Parkinson | 195 | 23 (23/0/0) | 2 |
| 10 | Titanic | 2201 | 3 (3/0/0) | 2 |
| 11 | Sonar | 208 | 60 (60/0/0) | 2 |
| Multi-class problems ||||||
| 12 | Iris | 150 | 4 (4/0/0) | 3 |
| 13 | Wine | 178 | 13 (13/0/0) | 3 |
| 14 | Balance | 625 | 4 (4/0/0) | 3 |
| 15 | Vehicle | 846 | 18 (0/18/0) | 4 |
| 16 | Nursery | 12960 | 8 (8/0/0) | 5 |

| 17 | Zoo | 101 | 16 (0/0/16) | 7 |
| 18 | Segment | 2310 | 19 (19/0/0) | 7 |
| 19 | Ecoli | 336 | 7 (7/0/0) | 8 |
| 20 | Pendigit | 10992 | 16 (0/16/0) | 10 |

* R: real, I: integer, N: nominal

We also examine the learners in this article on two real world data. The first dataset is German Traffic Sign Recognition Benchmark (GTSRB) [4]. The GTSRB images are collected from the roads in Germany during daytime and nighttime and the images have high variations due to the illumination, sunlight exposure, occluded by the obstacles of the roadsides, and rotations. There are in total more than 50,000 images; 39,209 for training and 12,630 for test. The number of classes of signs are 43 with highly unbalanced frequencies. The size of signs in the images has high variations. The second dataset is MNIST handwritten digit recognition database [41] with 70,000 images; 60,000 for training and 12,630 for test, in 10 classes. The images are black and white. The images were centered by calculating the center of mass of the pixels and the size for every image is fixed at 28 × 28. **Fig**. 2 displays instances for both MNIST and GTSRB data.

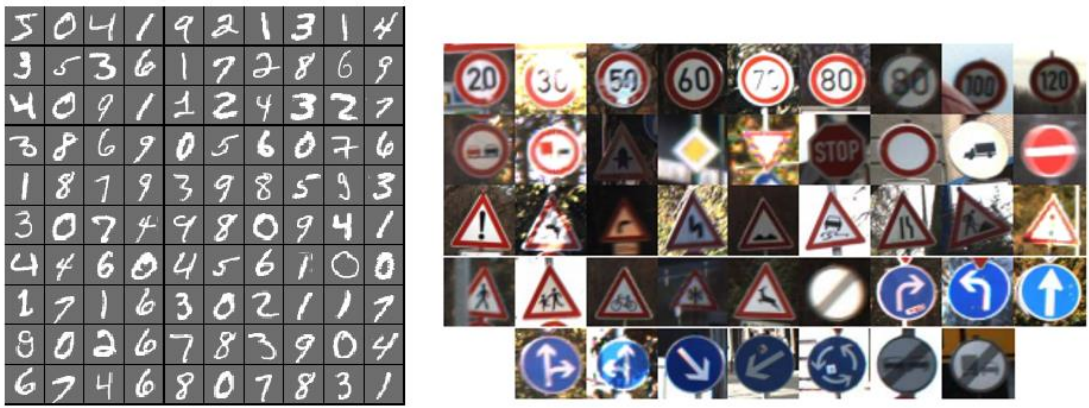

**Fig**. **2**. Examples of real-world datasets MNIST (in left) and GTSRB (in right)

To convert images to features, to be ready for learning purpose, we extract features with a histogram of Oriented Gradient (HOG) feature descriptor proposed by Dalal and Triggs at [42]. Cell size of HOG is set to 5 × 5. Every four cells construct a block and we consider 50% overlapping between blocks. The number of bins is set to 8. With these values assigned to HOG parameters, the total feature size is 1568.

### 4.2. Results on UCI Data

**Tables** 2 and 3 tabulate the average CA of all learners including the proposed *k*-RV on 20 UCI datasets with Gaussian and polynomial kernels respectively.

Following the results in **Tables** 2 and 3, *k*-RV is very competitive compared to other counterparts for both Gaussian and polynomial kernels. The second-best winner is RVM-Bern. The proposed *k*-RV is also a clear winner compared to ker-NN. *k*-RV outperforms ker-NN on several UCI datasets, with 4% gap in average CA for Gaussian kernel and 1.7% gap in for polynomial kernel. This implies that sparsity plays an important role in data classification with nearest neighbor rules.

**Table 2**. Average classification accuracy on 20 UCI data sets with Gaussian kernel.

| Index | Model / Data set | *k*-NN | *w*-NN | ker-NN | *k*-RV | RVM-Gauss | RVM-Bern |
|---|---|---|---|---|---|---|---|
| 1 | WBCD | 0.9011 | 0.9302 | 0.9267 | 0.9419 | 0.6964 | **0.9490** |
| 2 | Australia | 0.8845 | 0.9226 | 0.9191 | **0.9372** | 0.8519 | 0.8415 |
| 3 | Heart | 0.8702 | 0.8517 | 0.8626 | 0.8593 | **0.8633** | 0.8180 |
| 4 | Teaching | 0.6286 | 0.6492 | 0.6484 | **0.7978** | 0.7691 | 0.7572 |
| 5 | Iono | 0.8080 | 0.8055 | 0.8166 | **0.9105** | 0.8638 | 0.8502 |
| 6 | Pima | 0.7325 | 0.7392 | 0.7382 | **0.7771** | 0.7410 | 0.7441 |
| 7 | Bupa | 0.5611 | 0.5640 | 0.5522 | 0.6973 | **0.7164** | 0.6912 |
| 8 | Shuttle | 0.9047 | 0.9255 | 0.9252 | 0.9747 | 0.9817 | **0.9854** |
| 9 | Parkinson | 0.7703 | 0.7801 | 0.7844 | 0.8732 | **0.9349** | 0.8975 |
| 10 | Titanic | 0.8045 | **0.8059** | 0.8045 | 0.795 | 0.7703 | 0.7718 |
| 11 | Sonar | 0.8245 | 0.8207 | 0.8218 | **0.8476** | 0.6167 | 0.7833 |
| 12 | Iris | 0.90 | 0.9424 | 0.9333 | **0.9511** | 0.9347 | 0.9372 |
| 13 | Wine | 0.9050 | 0.9112 | 0.914 | **0.9963** | 0.9636 | 0.9657 |
| 14 | Balance | 0.8748 | 0.8805 | 0.8855 | **0.939** | 0.9371 | 0.9177 |
| 15 | Vehicle | 0.7505 | 0.7522 | **0.7548** | 0.7377 | 0.7317 | 0.7325 |
| 16 | Nursery | 0.9272 | 0.9320 | **0.9351** | 0.9275 | 0.9028 | 0.9112 |
| 17 | Zoo | 0.9433 | 0.9514 | 0.9503 | **0.9402** | 0.9139 | 0.9364 |
| 18 | Segment | 0.7422 | 0.7503 | 0.7496 | **0.9356** | 0.8396 | 0.8375 |
| 19 | Ecoli | 0.7840 | 0.7778 | 0.7919 | **0.8686** | 0.8547 | 0.8391 |
| 20 | Pendigit | 0.9901 | 0.9905 | 0.9928 | **0.9944** | 0.9818 | 0.9785 |
| Average | - | 0.8204 | 0.83 | 0.8311 | **0.8872** | 0.8429 | 0.8574 |

Bold values indicate the best value under same conditions.

**Table 3**. Average classification accuracy on 20 UCI data sets with polynomial kernel.

| Index | Model / Data set | *k*-NN | *w*-NN | ker-NN | *k*-RV | RVM-Gauss | RVM-Bern |
|---|---|---|---|---|---|---|---|
| 1 | WBCD | 0.9623 | 0.9715 | **0.9728** | 0.9695 | 0.7182 | 0.9222 |
| 2 | Australia | 0.7951 | 0.7880 | 0.7994 | **0.858** | 0.7826 | 0.7775 |
| 3 | Heart | 0.7722 | 0.7745 | 0.7782 | **0.8123** | 0.8001 | 0.7805 |
| 4 | Teaching | 0.5412 | 0.5603 | 0.5484 | 0.7978 | **0.7984** | 0.7428 |
| 5 | Iono | 0.8005 | 0.7914 | 0.7982 | 0.8194 | **0.8327** | 0.811 |
| 6 | Pima | 0.7720 | 0.7888 | 0.7993 | **0.839** | 0.7723 | 0.7711 |
| 7 | Bupa | 0.6658 | 0.6741 | **0.6926** | 0.6765 | 0.674 | 0.6625 |
| 8 | Shuttle | 0.9803 | 0.9858 | 0.9897 | **0.9947** | 0.9796 | 0.9805 |
| 9 | Parkinson | 0.9214 | 0.9345 | 0.9309 | **0.9542** | 0.9228 | 0.93 |
| 10 | Titanic | 0.7719 | 0.7806 | 0.7884 | 0.7909 | 0.8203 | **0.8217** |
| 11 | Sonar | 0.6605 | 0.6525 | 0.659 | 0.6986 | **0.7319** | 0.7283 |
| 12 | Iris | 0.9710 | 0.9803 | 0.9847 | **0.9984** | 0.8873 | 0.8755 |
| 13 | Wine | 0.9217 | 0.9335 | 0.9316 | **0.9585** | 0.9538 | 0.9412 |
| 14 | Balance | 0.8325 | 0.8480 | 0.8484 | **0.9075** | 0.8837 | 0.8718 |
| 15 | Vehicle | 0.8302 | **0.8325** | 0.8314 | 0.8210 | 0.8005 | 0.7932 |
| 16 | Nursery | 0.9340 | 0.9419 | 0.9435 | 0.9439 | **0.972** | 0.9538 |
| 17 | Zoo | 0.9015 | 0.8994 | 0.8903 | **0.9174** | 0.8329 | 0.7973 |
| 18 | Segment | 0.7127 | 0.7210 | 0.7296 | **0.7825** | 0.7327 | 0.7209 |
| 19 | Ecoli | 0.8749 | 0.8755 | 0.8761 | 0.8988 | **0.8996** | 0.8897 |
| 20 | Pendigit | 0.9715 | 0.9772 | 0.9810 | **0.9925** | 0.9705 | 0.9659 |
| Average | - | 0.8297 | 0.8356 | 0.8387 | **0.8678** | 0.8313 | 0.8319 |

Bold values indicate the best value under same conditions.

**Tables** 4 and 5 display the optimal number of RVs (# RV), the optimal width, the optimal $\Delta \alpha$ (delta) and the number of relevant attributes (#used) for Gaussian and polynomial kernels respectively. The main point to be mentioned, for all datasets, at least 95% of training vectors (#used) are pruned by *k*-RV while the performance of *k*-RV is still better than ker-NN.

**Table 4**. Number of used hidden nodes for several datasets (*Gaussian* hidden nodes)

| Dataset | *k*-RV | Dataset | *k*-RV |
|---|---|---|---|
| | (# RV, width, delta, # used) | | (# RV, width, delta, # used) |
| Wbcd | (6, 0.05 10, 0.012) | Iris | (1, 0.7, 0.1, 0.052) |
| Australia | (5, 0.4, 10, 0.035) | Wine | (2, 0.35, 5, 0.031) |
| Heart | (6, 5, 0.1, 0.12) | Balance | (1, 1, 5, 0.009) |
| Teaching | (6, 0.9, 0.1, 0.037) | Nursery | (41, 0.01, 1, 0.0032) |
| Iono | (1, 1, 5, 0.016) | Zoo | (6, 1, 1e-5, 0.06) |
| Pima | (8, 0.9, 5, 0.007) | Segment | (4, 0.5, 0.01, 0.04) |
| Bupa | (9, 0.95, 0.01, 0.026) | Ecoli | (5, 1, 1e-6, 0.017) |
| Shuttle | (1, 0.95, 5, 0.057) | Pendigit | (51, 0.1, 10, 0.0046) |
| Parkinson | (3, 0.75, 5, 0.017) | Vehicle | (2, 0.45, 1, 0.056) |
| Sonar | (6, 5, 0.1, 0.02) | Titanic | (7, 0.6, 5, 0.0025) |

**Table 5**. Number of used hidden nodes for several datasets (*polynomial* hidden nodes)

| Dataset | *k*-RV | Dataset | *k*-RV |
|---|---|---|---|
| | (# RV, width, delta, # used) | | (# RV, width, delta, # used) |
| Wbcd | (3, 0.25, 1e-6, 0.024) | Iris | (1, 0.6, 1e-5, 0.038) |
| Australia | (7, 0.4, 0.01, 0.003) | Wine | (5, 0.9, 10, 0.019) |
| Heart | (7, 1, 0.001, 0.078) | Balance | (1, 0.7, 0.001, 0.0036) |
| Teaching | (6, 0.9, 0.1, 0.037) | Nursery | (14, 0.45, 1e-6, 0.001) |
| Iono | (11, 1, 1e-5, 0.003) | Zoo | (6, 1, 1e-5, 0.06) |
| Pima | (5, 0.75, 1e-5, 0.013) | Segment | (4, 0.5, 0.01, 0.04) |
| Bupa | (9, 0.5, 1, 0.006) | Ecoli | (5, 1, 1e-6, 0.017) |
| Shuttle | (2, 0.85, 1e-6, 0.066) | Pendigit | (44, 0.1, 10, 0.0045) |

| Parkinson | (5, 0.5, 1e-4, 0.006) | Vehicle | (3, 0.5, 1, 0.059) |
| Sonar | (1, 0.6, 1e-4, 0.005) | Titanic | (3, 0.8, 0.001, 0.001) |

**Fig**. 3 displays the number of basis vectors and sparse basis vectors for ker-NN and $k$-RV respectively. The number of vectors in ker-NN is nearly 100 times greater than the non-zero vectors of $k$-RV for all UCI datasets.

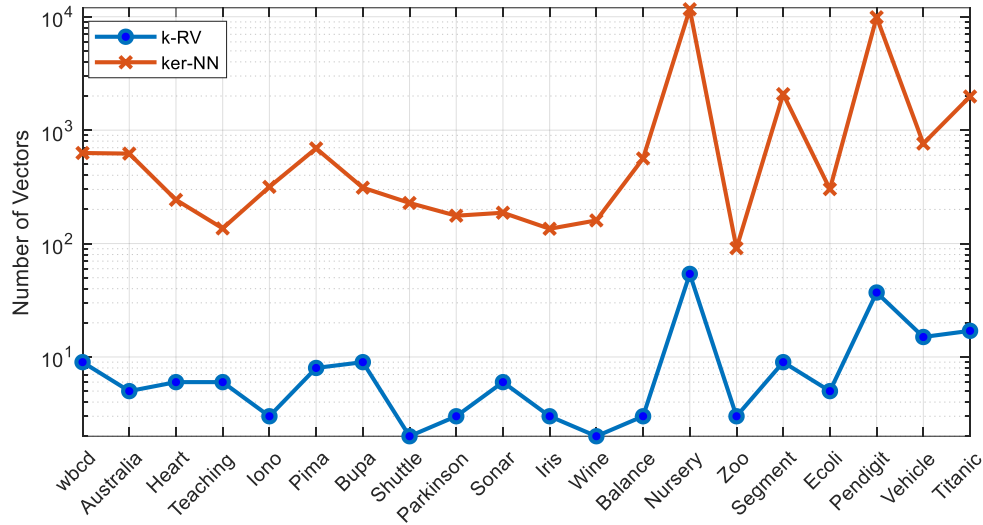

**Fig. 3** Training vectors of ker-NN versus relevance vectors of $k$-RV. The y-axis is in log scale

**Fig**. 4 illustrates relevant attributes and their weight values for Iris dataset. Only those vectors with non-zero weights take a part in the classification of query instances. The weight values are distributed in the range of [-1,1]. Among 120 attributes of training data, only 8 attributes and their corresponding non-zero weights are used to find $k$ relevance vectors through Eq. (8). This example emphasizes the ability of $k$-RV in significant sparsification of data while the CA is even better than a kernelized nearest neighbor rule that is not sparsed.

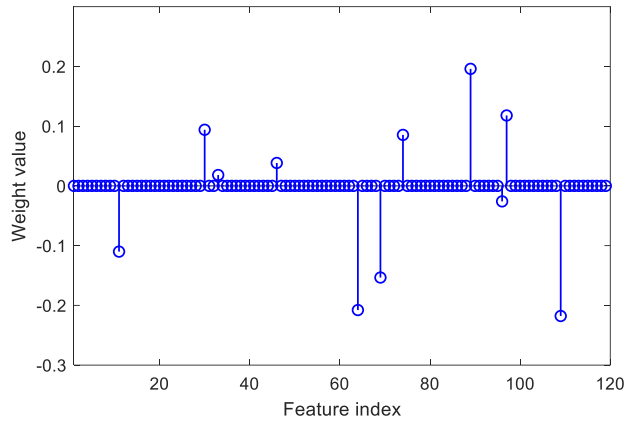

**Fig. 4.** Relevant attributes and their corresponding weight values

The *k*-RV and ker-NN models were run 10 times with 10-fold cross validation. Given 100 values of CAs, a statistical significance between *k*-RV and ker-NN based on a 95% confidence paired-t test is shown in **Table** 6. The hypothesis *S* with value '1' rejects the null hypothesis that the means of compared methods are equal, '0' otherwise.

**Table. 6.** 10 runs of 10-fold cross validation

| Index | Model / Data set | *k*-RV | ker-NN | S |
| --- | --- | --- | --- | --- |
| 1 | WBCD | 0.9419 | 0.9267 | 0 |
| 2 | Australia | **0.9372** | 0.9191 | 0 |
| 3 | Heart | 0.8593 | 0.8626 | 0 |
| 4 | Teaching | **0.7978** | 0.6484 | 1 |
| 5 | Iono | **0.9105** | 0.8166 | 1 |
| 6 | Pima | **0.7771** | 0.7382 | 1 |
| 7 | Bupa | 0.6573 | 0.5522 | 1 |
| 8 | Shuttle | 0.964 | 0.9252 | 1 |
| 9 | Parkinson | 0.8632 | 0.7844 | 1 |
| 10 | Titanic | 0.795 | **0.8045** | 0 |
| 11 | Sonar | 0.8476 | 0.8218 | 0 |
| 12 | Iris | 0.9511 | 0.9333 | 0 |
| 13 | Wine | **0.9963** | 0.914 | 1 |
| 14 | Balance | 0.929 | 0.8855 | 1 |
| 15 | Vehicle | 0.7177 | 0.7548 | 0 |
| 16 | Nursery | 0.9275 | 0.9351 | 0 |
| 17 | Zoo | **0.9402** | 0.9503 | 0 |
| 18 | Segment | **0.9356** | 0.7496 | 1 |
| 19 | Ecoli | **0.8686** | 0.7919 | 1 |
| 20 | Pendigit | **0.9944** | 0.9928 | 0 |

Bold values indicate the best value under same conditions.

Referring to **Table** 6, and the last column, the proposed *k*-RV is statistically significant compared to ker-NN on 10 out of 20 UCI datasets.

To further show the goodness of our proposed method, statistical significance test is performed for all learners with Friedman test. The Friedman test is a rank-based method that ranks each learner for each dataset. The best algorithm receives the rank 1, and the second best receives rank 2, and so on. Assume that $\rho_i^j$ is the rank of the $j^{th}$ of $G$ algorithms on the $i^{th}$ of $S$ datasets. The purpose is to compare the average rank of the algorithms $R_j$. The null hypothesis assumes that there is no difference between the algorithms. If the Fisher value $\pi_f$ is larger than the critical value $CV_\alpha$ acquired by the Fisher distribution with $G - 1$ and $(G - 1)(S - 1)$ degrees of freedom with $\alpha = 0.05$, the null hypothesis is rejected. The calculations for the Fisher value $\pi_f$ are (Demsar, 2006) as follows:

$$\chi_F^2 = \frac{12S}{G(G+1)} \left[ \sum_j R_j^2 - \frac{G(G+1)^2}{4} \right] \quad (9)$$

And,

$$\pi_f = \frac{(S-1)\chi_F^2}{S(G-1) - \chi_F^2} \quad (10)$$

The null-hypothesis is rejected if $\chi_F^2 > CV_\alpha$. If this happen, a Nemenyi diagram as post-hoc test (Demsar, 2006), is a good mean to visualize and compare the classifiers. Two classifiers are significantly different if their average ranks differ by at least the critical difference (CD). The CD can be obtained as follows.

$$CD = CV_\alpha \sqrt{\frac{G(G+1)}{6S}} \quad (11)$$

In our experiments, $CV_{0.05}$ for the two-tailed Nemenyi test is 11.0705 and $\chi_F^2 = 19.6571$. Because $\chi_F^2 > CV_\alpha$, the null hypothesis is rejected. The CD in our experiments is 1.6861. Obtained ranks for each learner is used to graphically compare the classifiers via the Nemenyi diagram as shown in **Fig. 5**.

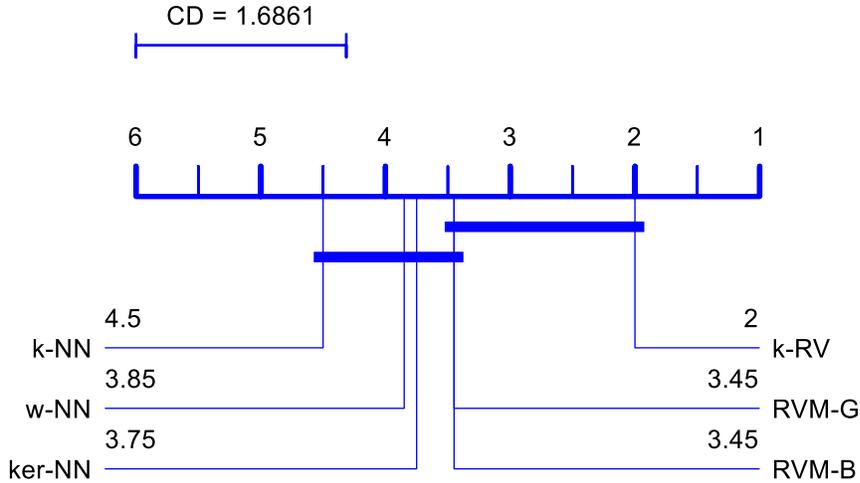

**Fig. 5.** Comparing the classifiers using the Nemenyi diagram

Referring to **Fig. 5**, the learners are categorized in two different sets (the thick horizontal bars) in terms of their ranks, i.e. first group = {*k*-RV, RVM-G and RVM-B} and second group = {*k*-NN, *w*-NN and ker-NN}. The proposed *k*-RV has best rank and *k*-NN has worst rank among all learners. The two classifiers have no significant difference when they are in the same group. Thus, the first group of learners, *k*-RV, RVM-G and RVM-B have no significant statistical difference. However, because the *k*-RV has better rank than RVM-G and RVM-B (2 compared to 3.45), *k*-RV is more statistically significant than all the other methods. It is worth noting that RVM-G and RVM-B are also in common with second group which means there is on statistical difference between them and k-NN, *w*-NN and ker-NN. But *k*-RV only belongs to the first group and hence is statistically significant than the second group. Applying SBL to *k*-NN rule, we devised a new nearest neighbor rule, called *k*-RV that its performance is statistically different from nearest neighbor rules, like *k*-NN, *w*-NN and ker-NN.

### 4.3 Computer Vision Data Experiments

Table 7 and Table 8 tabulate the performance of learners on two real-world datasets, GTSRB and MNIST, with Gaussian and polynomial kernels respectively.

**Table 7**. Classification accuracies and computed time on real-world datasets (*Gaussian* kernel)

| Model/Data set | GTSRB | | | MNIST | | |
|---|---|---|---|---|---|---|
| | Accuracy | Train time | Test time | Accuracy | Train Time | Test Time |

| | RVM-GK | 0.9406 | 30h 28min | 27sec | 0.9745 | 32h 22min | 31sec |
|---|---|---|---|---|---|---|---|
| | RVM-Bern | 0.9339 | 29h 10min | 29sec | 0.9719 | 32h 7min | 32sec |
| | ker-NN | 0.9217 | 4h 10min | 1min 43sec | 0.9606 | 5h 3min | 2min 28sec |
| | *k*-RV | **0.9526** | 53h 11min | 1min 14sec | **0.9841** | 57h 20min | 2min 7sec |

Boldface indicates the highest accuracy among classifiers under same conditions.

**Table 8**. Classification accuracies and computed time on real-world datasets (*polynomial* kernel)

| Model/Data set | GTSRB | | | MNIST | | |
|---|---|---|---|---|---|---|
| | Accuracy | Train time | Test time | Accuracy | Train Time | Test Time |
| RVM-Gauss | 0.9392 | 29h 5min | 28sec | 0.9722 | 33h 4min | 28sec |
| RVM-Bern | 0.9320 | 28h 2min | 26sec | 0.9705 | 30h 31min | 35sec |
| ker-NN | 0.9203 | 3h 44min | 1min 52sec | 0.9617 | 4h 52min | 2min 4sec |
| *k*-RV | **0.9488** | 50h 26min | 1min 10sec | **0.9825** | 55h 8min | 1min 42sec |

Boldface indicates the highest accuracy among classifiers under same conditions.

Comparing results in **Tables** 7 and 8, *k*-RV significantly outperforms ker-NN for both GTSRB and MNIST datasets, with polynomial and Gaussian kernel functions respectively. ker-NN is the fastest algorithm among all learners, however, it has the worst CA. *k*-RV with Gaussian kernel gains highest test accuracies 0.9526 and 0.9841 on GTSRB and MNIST respectively. The gap in test accuracies between *k*-RV and second-best winner, RVM-Gauss, is nearly 1%. However, this small gap means *k*-RV can correctly classify nearly 120 test images more than RVM-Gauss since GTSRB has more than 12,000 test images. But, the test time of RVM-Gauss is much better than our proposed *k*-RV since the test prediction in *k*-RV delays for every test instance as *k*-RV is a memory-based learner and regionally classifies each test instance. Moreover, referring to Eq. (6), RVM requires the inversion of covariance matrix that is timely. The test time of *k*-RV is better than ker-NN since k-RV uses sparsed weights and attributes to classify test instances while ker-NN is a dense learner and uses all weights and attributes.

## 5. Discussion

One aspect of our modelling to be mentioned is, RVM has poor predictions for those query instances are far away from the RVs. In [37], the source of problem is attributed to the degeneracy of covariance function. Intuitively, the fewer RVs, more probably the lower is the performance. Thus, the early stopping of RVM helps to have more RVs, hence better performance. we adjust the $\Delta\alpha$ and width of *Gaussian* kernel of basis functions to resolve the issue of the degeneracy of covariance function in RVM. **Fig**. 6 displays the plot of RVM application on Ripley's synthetic data. As it can be seen, adjustment of $\Delta\alpha$ generates more RVs. Distance is an important feature of our *k*-RV for accurate estimation of the target value and with generating more RVs, the input space has adjacent RVs for every query instance.

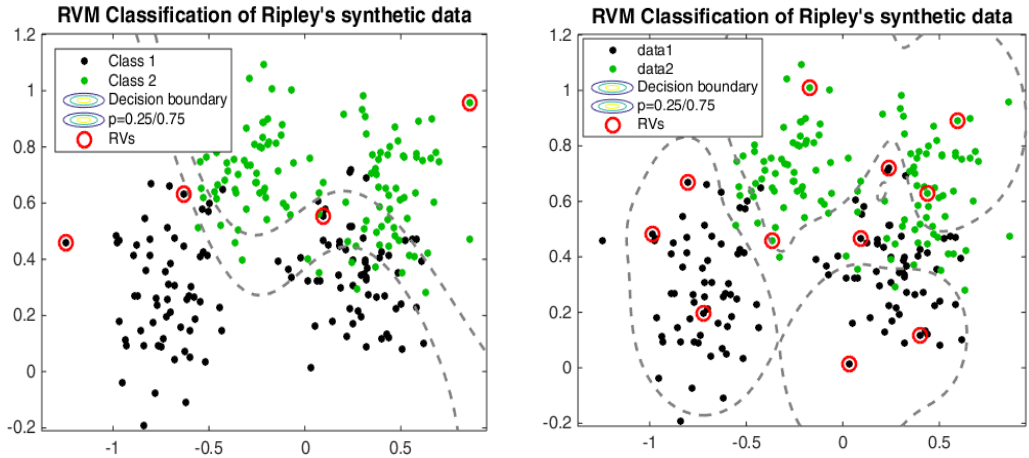

**Fig. 6**. The newly introduced parameter generates different number of RVs. Left: Relevance vectors of *k*-RV with $\Delta\alpha = 0.001$, Right: Relevance vectors of *k*-RV with $\Delta\alpha = 0.01$.

To verify the effectiveness of the new parameter $\Delta\alpha$, we also compare RVM-Gauss when $\alpha$ is fixed with RVM-Gauss when $\alpha$ is adjusted (RVM[+]-GK). **Table** 9 shows this comparison.

Table 9. Average classification accuracy of RVM-GK versus RVM[+]-GK

| Index | Model / Data set | RVM-Gauss | RVs | RVM[+]-Gauss | RV[+]s |
|---|---|---|---|---|---|
| 1 | WBCD | 0.6964 | 7 | 0.8875 | 19 |
| 2 | Australia | 0.8519 | 4 | 0.8827 | 9 |
| 3 | Heart | 0.8633 | 8 | 0.8694 | 18 |
| 4 | Teaching | 0.7691 | 9 | 0.7980 | 22 |
| 5 | Iono | 0.8638 | 3 | 0.8953 | 8 |
| 6 | Pima | 0.741 | 6 | 0.8125 | 14 |
| 7 | Bupa | 0.7164 | 10 | 0.7449 | 20 |
| 8 | Shuttle | 0.9817 | 4 | 0.9822 | 13 |
| 9 | Parkinson | 0.9349 | 5 | 0.9579 | 14 |
| 10 | Titanic | 0.7703 | 5 | 0.7993 | 11 |
| 11 | Sonar | 0.6167 | 5 | 0.6645 | 19 |
| 12 | Iris | 0.9347 | 3 | 0.9490 | 14 |
| 13 | Wine | 0.9636 | 3 | 0.9632 | 11 |
| 14 | Balance | 0.9571 | 7 | 0.9574 | 26 |
| 15 | Vehicle | 0.7317 | 8 | 0.7433 | 14 |
| 16 | Nursery | 0.9028 | 40 | 0.9155 | 48 |
| 17 | Zoo | 0.9139 | 7 | 0.9373 | 11 |
| 18 | Segment | 0.8396 | 5 | 0.8522 | 42 |
| 19 | Ecoli | 0.8547 | 4 | 0.8522 | 14 |
| 20 | Pendigit | 0.9818 | 54 | 0.9836 | 95 |
| Average | - | 0.8429 | 9.85 | **0.8727** | 22.1 |

Referring to **Table** 9, RVM$^+$-Gauss has better CA than RVM-Gauss on most datasets. To justify this better performance, we observe that the number of RV$^+$s is greater than the number of RVs for every dataset. This implies that the traditional RVM prunes the datasets excessively, resulting in the degeneracy of covariance function [37]. The degeneracy means that the rank of the kernel matrix is equal to the number of RVs. Thus, the matrix is very low rank and ill-conditioned. The new parameter $\Delta\alpha$, plays as a role of early stopping. This parameter does not allow RVM to prune data undesirably. Hence, more RVs are preserved and the rank increases. **Fig**. 6 also shows how adjusting the new parameter $\Delta\alpha$ generates more RVs.

## 6. Conclusion

In this study, we linked two different learning approaches, memory-based learning and statistical learning. We showed that only very few relevant attributes and their corresponding weights are enough to improve the classification accuracy of the *k*-NN. Experiments conducted on several datasets with two different types of kernel showed the advantage of the proposed *k*-RV over a few state-of-the-arts. The new parameter that controls the stopping condition of RVM helped to produce more RVs and heal the degeneracy of covariance function in RVM. Thus, higher classification accuracy obtained by *k*-RV. Future research direction is conducted on improving the speed of the proposed algorithm with graphical processing unit (GPU) since k-RV has slow training time.

**Author Contribution**

Peyman Hosseinzadeh Kassani and Sara Hosseinzadeh Kassani are co-first authors and contributed equally to this work.